\documentclass[10pt,twocolumn,letterpaper]{article}

\usepackage{cvpr}
\usepackage{epsfig}
\usepackage{booktabs}
\usepackage{times,graphicx,tabulary,multirow,xspace}
\usepackage[dvipsnames]{xcolor}
\usepackage{amsmath,amssymb,soul,setspace,pifont}
\usepackage{makecell}
\usepackage[caption=false]{subfig}
\usepackage[british,american]{babel}
\usepackage{url}
\usepackage[font=footnotesize,labelfont=bf]{caption}
\usepackage{appendix}
\usepackage[symbol]{footmisc}

\usepackage[pagebackref=true,breaklinks=true,letterpaper=true,colorlinks,
  citecolor=citecolor,bookmarks=false]{hyperref}
\definecolor{citecolor}{RGB}{34,139,34}

\newcommand{\graytext}[1]{\textcolor{gray}{#1}}
\newcommand{\trainset}[0]{\texttt{trainval}\xspace}
\newcommand{\trainonly}[0]{\texttt{train}\xspace}
\newcommand{\valminusminival}[0]{\texttt{vqa-eval}\xspace}
\newcommand{\minival}[0]{\texttt{vqa-dev}\xspace}
\newcommand{\testset}[0]{\texttt{test-dev}\xspace}
\newcommand{\testfinal}[0]{\texttt{test-std}\xspace}
\newcommand{\butd}[0]{\textsc{R}\xspace}
\newcommand{\ours}[0]{\textsc{G}\xspace}
\definecolor{mygray}{gray}{0.4}
\newcommand{\cmark}{\color{mygray}\ding{51}}%
\newcommand{\xmark}{\color{mygray}\ding{55}}%
\newcolumntype{x}[1]{>{\centering\arraybackslash}p{#1pt}}
\newlength\savewidth\newcommand\shline{\noalign{\global\savewidth\arrayrulewidth
  \global\arrayrulewidth 1pt}\hline\noalign{\global\arrayrulewidth\savewidth}}
\newcommand\hshline{\noalign{\global\savewidth\arrayrulewidth
  \global\arrayrulewidth 0.6pt}\hline\noalign{\global\arrayrulewidth\savewidth}}
\newcommand{\tablestyle}[2]{\setlength{\tabcolsep}{#1}\renewcommand{\arraystretch}{#2}\centering\footnotesize}
\makeatletter\renewcommand\paragraph{\@startsection{paragraph}{4}{\z@}
  {.5em \@plus1ex \@minus.2ex}{-.5em}{\normalfont\normalsize\bfseries}}\makeatother
\def\x{$\times$\@\xspace}
\newcommand{\app}{\raise.17ex\hbox{$\scriptstyle\sim$}}

\newcommand{\roi}{\texttt{RoIPool}\xspace}
\newcommand{\fc}{\texttt{FC}\xspace}

\cvprfinalcopy %

\def\cvprPaperID{1175} %

\ifcvprfinal\fi
\begin{document}

\title{In Defense of Grid Features for Visual Question Answering}

\renewcommand*{\thefootnote}{\fnsymbol{footnote}}
\author{Huaizu Jiang$^{1,2}$\footnotemark~, Ishan Misra$^2$, Marcus Rohrbach$^2$, Erik Learned-Miller$^1$, and Xinlei Chen$^2$  \\
$^1$UMass Amherst, $^2$Facebook AI Research (FAIR) \\
{\tt\small \{hzjiang,elm\}@cs.umass.edu}, {\tt\small \{imisra,mrf,xinleic\}@fb.com}
}

\maketitle
\thispagestyle{empty}

\begin{abstract}
   Popularized as `bottom-up' attention~\cite{anderson2018bottom}, bounding box (or region) based visual features have recently surpassed vanilla grid-based convolutional features as the de facto standard for vision and language tasks like visual question answering (VQA). However, it is not clear whether the advantages of regions (\eg better localization) are the key reasons for the success of bottom-up attention. In this paper, we revisit grid features for VQA, and find they can work surprisingly well -- running more than an order of magnitude faster with the same accuracy (\eg if pre-trained in a similar fashion). Through extensive experiments, we verify that this observation holds true across different VQA models (reporting a state-of-the-art accuracy on VQA 2.0 \testfinal, \textbf{72.71}), datasets, and generalizes well to other tasks like image captioning. As grid features make the model design and training process much simpler, this enables us to train them end-to-end and also use a more flexible network design. We learn VQA models end-to-end, from pixels directly to answers, and show that strong performance is achievable without using any region annotations in pre-training. We hope our findings help further improve the scientific understanding and the practical application of VQA. Code and features will be made available. 
\end{abstract}

\footnotetext[1]{This work was done when Huaizu Jiang was an intern at FAIR.}
\renewcommand*{\thefootnote}{\arabic{footnote}}
\setcounter{footnote}{0}

\section{Introduction}

After the introduction of deep learning~\cite{donahue2015long,vinyals2015show} and attention mechanisms~\cite{xu2015show,yang2016stacked} to multi-modal vision and language research, perhaps one of the most significant developments was the discovery of `bottom-up' attention~\cite{anderson2018bottom}. Unlike normal attention that uses `top-down' linguistic inputs to focus on specific parts of the visual input, bottom-up attention uses pre-trained object detectors~\cite{ren2015faster} to identify salient regions based \emph{solely} on the visual input itself. As a result, images are represented by a collection of bounding box or \textbf{region}\footnote{We use the terms `region' and `bounding box' interchangeably.}-based features~\cite{anderson2018bottom,teney2018tips}--in contrast to vanilla \textbf{grid} convolutional feature maps from ConvNets~\cite{simonyan2014very,he2016deep}--for follow-up tasks. These region features have since then gained wide popularity and dominated vision and language leader boards~\cite{jiang2018pythia,yu2019deep} for major tasks like visual question answering (VQA).

So what makes these region features successful? Naturally, one would assume a major reason is better \emph{localization} of individual objects, as the regions are direct bounding box outputs from detectors. Another plausible answer is that a number of regions can easily capture both the coarse-level information and fine-grained details in the image -- even if they overlap. However, do these potential advantages actually demonstrate that region features are superior to grids?

\begin{figure}[t]
\centering
\includegraphics[width=1\linewidth]{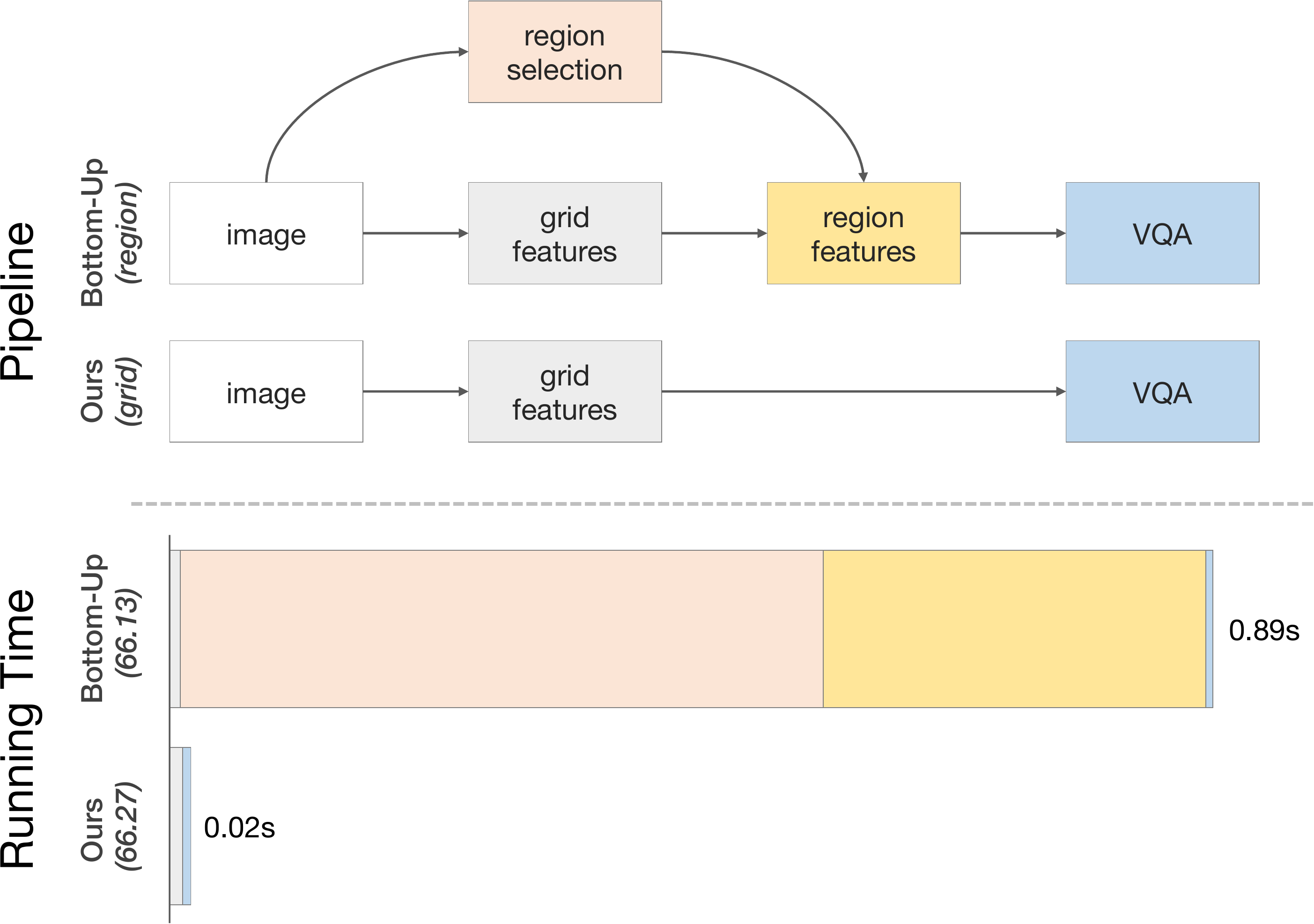}
\caption{We revisit {\bf grid}-based convolutional features for VQA, and find they can \emph{match} the accuracy of the dominant {\bf region}-based features from bottom-up attention~\cite{anderson2018bottom}, provided that one closely follow the pre-training process on Visual Genome~\cite{krishna2017visual}. As computing grid features skips the expensive region-related steps (shown in colors), it leads to significant speed-ups (all modules run on GPU; timed in the same environment).}
\label{fig:teaser}\vspace{-2mm}
\end{figure}

Surprisingly, we discovered that grid features extracted from \emph{exactly} the same layer of the pre-trained detector can perform competitively against their region-based counterparts for VQA. Moreover, with simple modifications during training, the same grid features can be made even more effective and that they consistently achieve comparable and sometimes better VQA accuracy than region features. In fact, our ablative analysis suggests that the key factors which contributed to the high accuracy of existing bottom-up attention features are: 1)~the large-scale object and attribute annotations collected in the Visual Genome (VG)~\cite{krishna2017visual} dataset used for pre-training; and 2)~the high spatial resolution of the input images used for computing features. As for the feature \emph{format} itself -- region or grid -- it only affects accuracy \emph{minimally}. Through a comprehensive set of experiments, we verified that our observations generalize across different network backbones, different VQA models~\cite{jiang2018pythia,yu2019deep}, different VQA benchmarks~\cite{antol2015vqa,gurari2018vizwiz}, and even to other relevant tasks (\eg image captioning~\cite{chen2015microsoft}). 

Our findings have important consequences for the design of future multi-modal vision and language models. The immediate benefit of switching to grids is inference speed, as we can now skip \emph{all} of the region-related steps in the existing VQA pipeline (Fig.~\ref{fig:teaser}). For example, using a ResNet-50~\cite{he2016deep} backbone, we find the overall running time drops from 0.89s to 0.02s per image -- \textbf{40+} times faster with slightly better accuracy! In fact, extracting region features is so time-consuming that most state-of-the-art models~\cite{kim2018bilinear,yu2019deep} are directly trained and evaluated on \emph{cached} visual features. This practice not only imposes unnecessary constraints on model designs, but also limits potential applications of existing vision and language systems. 

Empowered by grid features, we therefore take an initial step to train VQA models \emph{end-to-end} from pixels directly to answers. Note that end-to-end training with region features is challenging, since fine-tuning region locations likely requires additional grounding annotations~\cite{harnad1990symbol} that are computationally expensive and difficult to acquire. In contrast, grid features can be readily optimized for the final objective (\eg to answer questions correctly) without extra grounding. The grid-feature pipeline also allows us to explore more effective designs for VQA (\eg pyramid pooling module~\cite{zhao2017pyramid}) and enables networks pre-trained with \textbf{zero} region-level annotations to greatly reduce the gap in accuracy with VG models (trained on bounding boxes) -- indicating strong VQA models can be achieved without \emph{any} explicit notion of regions. These results further strengthen our defense of grid features for VQA. We hope our discovery can open up new opportunities for vision and language research in general.

\section{Related Work}
\label{sec:related_work}
\paragraph{Visual features for vision and language tasks.}  Features have played a key role in the advancement of vision and language tasks. For example, deep learning features led to remarkable improvements in image captioning~\cite{donahue2015long,vinyals2015show,devlin2015exploring}. While a complete review of visual features used for vision and language tasks is beyond the scope of this paper, we note that the accuracies of modern VQA models are dependent on the underlying visual features used, including VGG~\cite{simonyan2014very} and ResNet~\cite{he2016deep} grid features, which were later dominated by bottom-up attention region features~\cite{anderson2018bottom,teney2018tips}. Today, most state-of-the-art VQA models focus on fusing schemes~\cite{yu2017multi,kim2018bilinear,yu2019deep} and are built with region features as-is~\cite{yi2018neural}; whereas our work revisits grid features, and shows that they can be equally effective and lead to remarkable speed-ups -- often greater than an order of magnitude!

\par \noindent \textbf{Pre-training for VQA.} Most VQA methods use two separately pre-trained models: vision models trained on ImageNet~\cite{deng2009imagenet} and VG~\cite{krishna2017visual}; and word embeddings~\cite{pennington2014glove} for linguistic features. As these separately trained features may not be optimal for joint vision and language understanding, a recent hot topic is to develop jointly pre-trained models~\cite{li2019visualbert,lu2019vilbert,tan2019lxmert,su2019vl,zhou2019unified,chen2019uniter} for vision and language tasks. A common scheme for such methods is to view regions and words as `tokens' for their respective domain, and pre-train a variant of BERT~\cite{devlin2018bert,vaswani2017attention} for `masked' token prediction. Complementary to that direction, our work delves specifically into the `format' of visual tokens and can be potentially combined with such methods for mutual benefits (\eg trade-off between speed and accuracy).

\par \noindent \textbf{Regions \vs grids.} The debate between region features and grid features carries some inherent connections to object detection: the dominance of the R-CNN based detection models~\cite{ren2015faster,He2017} demonstrates that a region (the `R' in R-CNN) based refinement stage is beneficial for object detection. On the other hand, one-stage detectors~\cite{lin2017focal,liu2016ssd} approach the detection task \emph{without} the need for explicit region-level computation and show that grid features can be competitive for object detection. In our work, we also use grid features -- \emph{no} regions for the VQA task. To minimize changes from bottom-up attention paper~\cite{anderson2018bottom}, we pre-train the features with Faster R-CNN~\cite{ren2015faster}. However, during inference, we discard the region-related steps from the detector and use \emph{only} the grid convolutional features. This in fact gives us a \emph{stronger} defense for grids, as we show that VQA can operate on a `single' feature map, instead of feature maps of `multiple' scales that one-stage detectors~\cite{lin2017focal,liu2016ssd} thrive on. 

\begin{figure*}[t]
\centering
\includegraphics[width=1\linewidth]{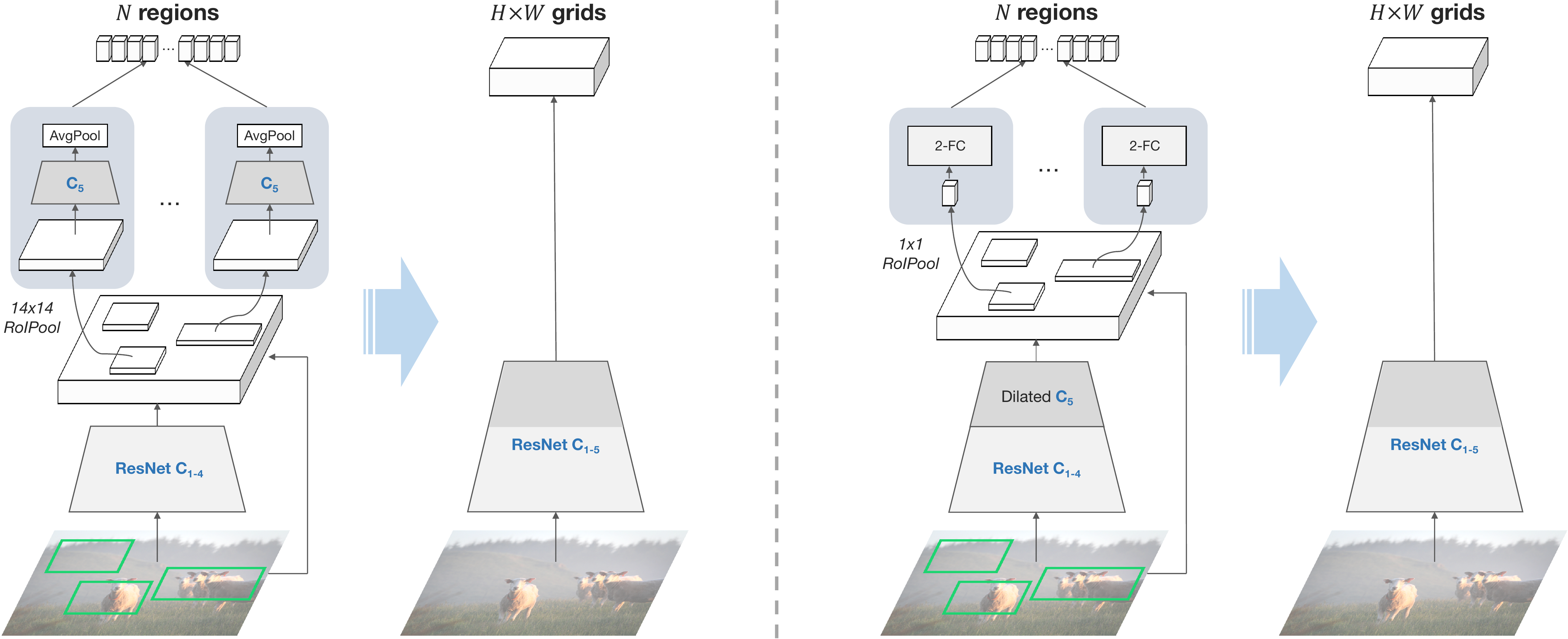}
\caption{\textbf{From regions to grids.} \textbf{Left}: We convert the original region feature extractor used by bottom-up attention~\cite{anderson2018bottom} back to the ResNet~\cite{he2016deep} grid feature extractor for the \emph{same} layer (see Sec.~\ref{sub_sec:same_layer}, weights in {\color{RoyalBlue} blue} are transferred), and find it works surprisingly well for VQA~\cite{goyal2017making}. \textbf{Right}: We build a detector based on 1{\x}1 \roi while keeping the output architecture \emph{fixed} for grid features (see Sec.~\ref{sub_sec:improved_grids}), and the resulting grid features consistently perform at-par with region features.}
\label{fig:approach}
\end{figure*}

It is also worth noting that while region features are effective on benchmarks like VQA~\cite{antol2015vqa,goyal2017making} and COCO captions~\cite{chen2015microsoft}, for benchmarks that diagnose a model's reasoning abilities when answering visual questions (\eg CLEVR~\cite{johnson2017clevr}), simple methods based on grids~\cite{perez2018film} have shown strong performance. We hope that our discovery that grid features also work well for the general VQA task can bridge the gap between these two lines of work~\cite{shrestha2019answer}.

\section{From Regions to Grids}
\label{sec:methodology}
In this section, we explain our approach to obtaining grid features that are just as effective as region features, with the constraint that they have been pre-trained with the \emph{same} task. In Sec.~\ref{sec:e2e_vqa}, we show that the `same pre-training' constraint can be lifted and grid features can still close the gap to regions with end-to-end training on down-stream tasks. We first briefly review the region features from bottom-up attention~\cite{anderson2018bottom}.

\subsection{Bottom-Up Attention with Regions}
\label{sub_sec:butd}
The bottom-up attention method~\cite{anderson2018bottom} uses a Faster R-CNN~\cite{ren2015faster} detection model. The detector is trained on a cleaned version of Visual Genome~\cite{krishna2017visual}, with thousands of object categories and hundreds of attributes with bounding box (region) annotations.

In order to obtain bottom-up attention features for tasks like VQA, two region-related steps are needed: 
\par \noindent \textbf{Region selection.} As Faster R-CNN is a two-stage detector, region selection happens twice in the pipeline. The first is through a region proposal network~\cite{ren2015faster}, which deforms and selects prominent candidate `anchors' as Regions of Interest (RoIs). Another selection is done as post-processing to aggregate top $N$ boxes in a \emph{per-class} manner. In both steps, non-maximal suppression (NMS) is used, which keeps the region with the highest classification score and removes other near-duplicates in a local neighborhood. 

\par \noindent \textbf{Region feature computation.} Given regions from the first stage (up to thousands), \roi operations~\cite{ren2015faster} are used to extract the initial region-level features. Additional network layers then compute the output representation of regions \emph{separately}. Finally, region features that survive both rounds of selection are stacked together as the bottom-up features to represent an image.

It is important to note that due to the complexity of the VG dataset (\eg thousands of classes) and the specific Faster R-CNN detector used~\cite{anderson2018bottom} (described next), both steps are computationally intensive. In contrast, directly using grid features can skip or accelerate these steps and offer potentially significant speed-ups.

\subsection{Grid Features from the Same Layer}
\label{sub_sec:same_layer}
The simplest way to convert region features to grids is to see if one can directly compute outputs of the same network layer, but in a \emph{shared}, fully convolutional manner. To this end, we take a closer look at the specific Faster R-CNN architecture used by the original bottom-up attention~\cite{anderson2018bottom}. 

The Faster R-CNN is a variant of the \emph{c4} model~\cite{he2016deep} with an extra branch for attribute classification. It divides the weights from a ResNet~\cite{he2016deep} into two separate sets: given an input image, it first computes feature maps using the lower blocks of ResNet up to $C_4$. This feature map is shared among all regions. Then, separately, per-region feature computations are performed by applying the $C_5$ block on the 14{\x}14 \roi-ed features. The output of $C_5$ is then \emph{AvgPool}-ed to a final vector for each region as the bottom-up features~\cite{anderson2018bottom}. Since all the final region features are from $C_5$, it is easy to convert the detector \emph{back} to the ResNet classifier and take the same $C_5$ layer as our output grid features. Fig.~\ref{fig:approach} (left) illustrates our conversion process.

As our experiments will show, directly using the converted $C_5$ output already works surprisingly well. Any performance drop from doing so may be because Faster R-CNN is highly optimized for \emph{region}-based object detection, and likely not so much for grids. Therefore, we next see if some minimal adjustments to the model can be made to improve grid features.

\subsection{1{\x}1 {\bf\roi} for Improved Grid Features}
\label{sub_sec:improved_grids}
Our idea is to simply use 1{\x}1 \roi. This means representing each region with a single \emph{vector}, rather than a three-dimensional tensor in  Faster R-CNN. At first glance, it may seem counter-intuitive, as the two additional spatial dimensions (height and width) are useful to characterize different parts of objects in 2D -- indeed, we find this modification negatively affects object detection performance on VG. But importantly, using 1{\x}1 \roi regions also means each vector on the grid feature map is \emph{forced} to cover all the information for a spatial region \emph{alone}, which can potentially result in stronger grid features.

However, directly applying 1{\x}1 \roi on the original model is problematic, likely because $C_5$ consists of several ImageNet pre-trained convolutional layers that work best with inputs of particular spatial dimensions. To resolve this, we follow recent developments in object detection and use the entire ResNet up to $C_5$ as the backbone for shared feature computation~\cite{zhu2019deformable}; and for region-level computation place two 1024D fully-connected (\fc) layers on the top, which by \emph{default} accept vectors as inputs. 

To reduce the effect of low resolutions when training the detector with features pooled from $C_5$ ($C_5$ has stride 32, whereas $C_4$ has 16), the stride-2 layers are replaced with stride-1 layers, and the remaining layers are dilated with a factor of 2~\cite{zhu2019deformable}. For grid feature extraction, we remove this dilation and convert it back to the normal ResNet.

Fig.~\ref{fig:approach} (right) summarizes the changes we made to improved grids. Note that compared to the original model (left), we only made necessary modifications to the region related components during training. Since all such computations are removed during feature extraction, our grid feature extractor is kept \emph{untouched} during inference.

\newlength{\Oldarrayrulewidth}
\newcommand{\scline}[1]{%
  \noalign{\global\setlength{\Oldarrayrulewidth}{\arrayrulewidth}}%
  \noalign{\global\setlength{\arrayrulewidth}{1pt}}\cline{#1}%
  \noalign{\global\setlength{\arrayrulewidth}{\Oldarrayrulewidth}}}
  
\newcommand{\rownumber}[1]{\textcolor{Cerulean}{#1}}
  
\begin{table}[t]
\tablestyle{5pt}{1.2}
\centering
\begin{tabular}{c  c|c|c|c|c|c}
 & &\multicolumn{3}{c|}{VG detection pre-train} & \multicolumn{2}{c}{VQA} \\
  \scline{2-7}
\rownumber{\#} & feature &  \roi & region layers & AP & accuracy & $\Delta$ \\
\cline{2-7}
\rownumber{1} & \multirow{2}{*}{\butd~\cite{anderson2018bottom}} & 14{\x}14 & $C_5$~\cite{he2016deep} & 4.07 & \underline{64.29} & - \\
\rownumber{2} & & 1{\x}1 & 2-\fc & 2.90 & 63.94 & \emph{-0.35} \\
\cline{2-7}
\rownumber{3} & \multirow{3}{*}{\ours} & 14{\x}14 & $C_5$ & 4.07 & 63.64 & \emph{-0.65} \\
\rownumber{4} & & 1{\x}1 & 2-\fc & 2.90 & \textbf{64.37} & \emph{0.08} \\
  \cline{3-7}
\rownumber{5} & & \multicolumn{3}{c|}{\graytext{ImageNet pre-train}} & \graytext{60.76} & \graytext{\it -3.53} \\
\end{tabular}
\caption{\textbf{Main comparison}. `\butd' stands for region features as in bottom-up attention~\cite{anderson2018bottom}. `\ours' stands for grid features. All results reported on VQA 2.0 \valminusminival. We show that: \textbf{1)} by simply extracting grid features from the \emph{same} layer $C_5$ of the same model, the VQA accuracy is already much closer to bottom-up attention than ImageNet pre-trained ones (row 1,3 \& 5); \textbf{2)} 1{\x}1 \roi based detector pre-training improves the grid features accuracy while the region features get worse (row 1,2 \& 4). Last column is the gap compared to the original bottom-up features (underlined).}
\label{tab:dc5_vs_c4}
\vspace{-4mm}
\end{table}

\section{Main Comparison: Regions \vs Grids}
\label{sec:main_results}
From this section on, we report our experimental results comparing regions with grids. We choose VQA (2.0)~\cite{goyal2017making} as our main task of interest, since it is currently a major benchmark for evaluating joint vision and language understanding and has clear metrics for evaluation.
For all our comparisons, we denote methods using region features with the tag `\butd', and methods using grid features with `\ours'. In this section, we focus on reporting our main findings from converting regions to grids as described in Sec.~\ref{sec:methodology}. We begin by briefly describing our experimental setups (more details in the supplementary material). Note that our goal here is to make the conclusion \emph{meaningful} by controlled comparisons, and not necessarily to optimize for absolute performance. 

\subsection{Experimental Setup}
\label{subsec:main_results_expt_setup}
\par \noindent \textbf{Faster R-CNN.} For analysis, we use Faster R-CNN with a ResNet-50 backbone pre-trained on ImageNet by default\footnote{\url{https://github.com/facebookresearch/maskrcnn-benchmark}}. Closely following bottom-up attention~\cite{anderson2018bottom}, the detector is then trained on the VG dataset~\cite{krishna2017visual} with region-level annotations for 1600 object categories and 400 attribute classes. For attributes, an additional branch is added with loss weight 0.5. The model is trained with `1x' schedule~\cite{He2017}. Notably, input images are resized to have a maximum shorter side of 600 pixels (longest 1000) when keeping aspect ratio fixed. For region features, we set $N$=100.

\par \noindent \textbf{VQA split.} Unless otherwise specified, we use the default \trainonly set for training. To assist our analysis, we create a local validation set, \minival, out of the standard \texttt{val} set to select the best model during training for evaluation. It contains randomly sampled 8.4K images and their corresponding questions, with 66K pairs in total. The rest of the original \texttt{val} set (named \valminusminival) is reserved for testing, on which we report results.

\par \noindent \textbf{VQA model.} We use the co-attention model~\cite{yu2018beyond} implemented in Pythia~\cite{jiang2018pythia,singh2019TowardsVM}. This model fuses visual features (either region or grid) with textual representations of questions, and outputs the final answer. 

\subsection{Main Results}
\label{sec:1x1_roi_result}
Our main results are summarized in Table~\ref{tab:dc5_vs_c4}. We make two observations: 
First, compared with the widely used bottom-up region features (row 1), directly extracting outputs from $C_5$ with the same model (row 3) works \emph{surprisingly} well (64.29 \vs 63.64 accuracy). In contrast, the standard ResNet-50 model pre-trained on ImageNet~\cite{deng2009imagenet} shows much worse performance -- 60.76 accuracy, a gap of more than 3\% with the bottom-up features.

\begin{figure}[t]
\centering
\includegraphics[width=0.95\linewidth]{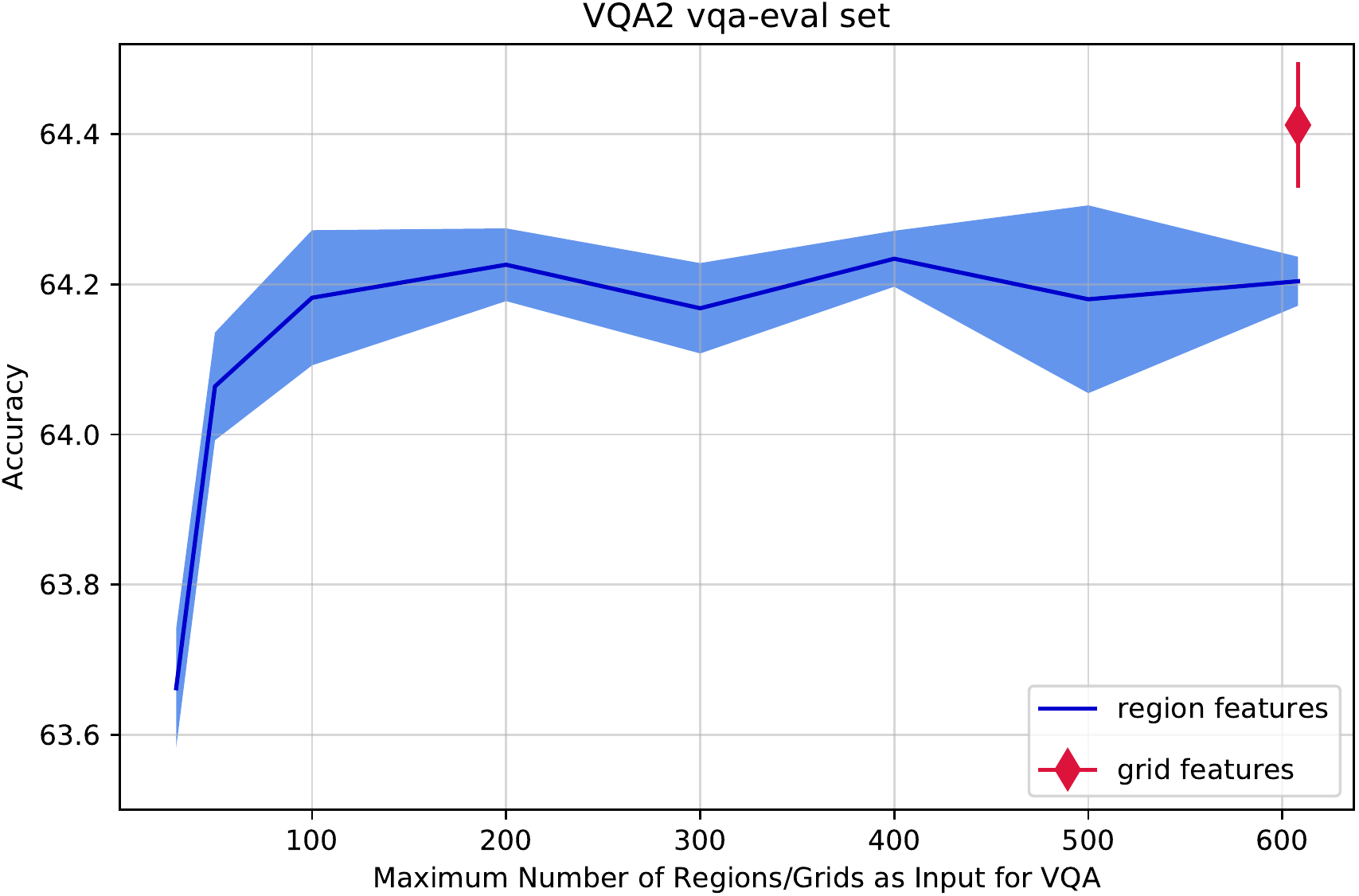}
\caption{\textbf{VQA accuracy \vs number of features $N$} as input to the VQA model. We report the average accuracy and standard deviation across 5 independent runs on the VQA 2.0 \valminusminival set. We observe that the VQA accuracy of region features saturates around 200 regions. In contrast, the grid features benefit from a larger $N$ (translates from a larger input size) and in this case stays better than regions even when $N$ is the same (608).
}
\label{fig:num_tokens}
\vspace{-2mm}
\end{figure}

Second, while our 1{\x}1 \roi-based variant hurts the object detection performance (average precision~\cite{lin2014microsoft} on VG drops from 4.07 to 2.90), it helps VQA -- boosting the accuracy by 0.73\% (row 3 \& 4) and as a result slightly \emph{outperforms} the original region-based features. On the other hand, our RoI-based variant does not help the region features method and drops the accuracy of region features to 63.94. This indicates the original model used by bottom-up attention favors regions; while our design works better for grids. Thus, we use the setting of the 1$^{st}$ row (best for regions) to represent `\butd', and the 4$^{th}$ row (best for grids) to represent `\ours', to perform a more in-depth study and fair comparison between the two through the rest of the paper.

\subsection{Number of Regions}
Apart from architectural differences in training, another factor that can affect VQA accuracy is the number of feature vectors $N$ used to represent images. 
Our region model from Pythia~\cite{jiang2018pythia} has a default setting that uses the top 100 boxes to represent region features, increasing it from the original 36 boxes in~\cite{anderson2018bottom} to improve the accuracy. On the other hand, since grid features are convolutional feature maps for a pre-set layer, the number of features is determined by the input size to the network. As our largest input size is 600{\x}1000, a 32-stride feature map ($C_{5}$) results in 608 grid features -- much larger than the number of region features. To understand how these different numbers of region features affect the accuracy, we ran experiments with varying number of features $N$ and show the results in Figure~\ref{fig:num_tokens}.

As for the region features, we observe an improvement in accuracy as the number of regions increases from 30 to 200, beyond which the accuracy saturates. Interestingly, our grid features are better even when compared to the highest number of regions\footnote{Since NMS is used in selecting regions, the maximum number $N$ varies across images. Therefore we 1) cannot directly set it to the same number as grids and 2) report maximum $N$ instead (zero paddings are used for images with fewer regions).}. Thus, the higher number of feature vectors used in our grid method compared to the baseline region method, is not the reason for its improved VQA accuracy.

\begin{table}
\tablestyle{2pt}{1.2}
\begin{tabular}{c|c|c|c|c|c|c|c}
 & \multirow{3}{*}{\makecell{\# features \\ ($N$)}} & \multirow{3}{*}{\makecell{\testset\\accuracy}} & \multicolumn{5}{c}{inference time breakdown (ms)} \\
 \cline{4-8}
 & & & \makecell{shared\\conv.} & \makecell{region\\feat. comp.} & \makecell{region\\selection} & VQA & total \\
\shline
\multirow{2}{*}{\butd} & 100 & 66.13 & 9 & 326 & 548 & 6 & 889 \\\
& 608 & 66.22 & 9 & 322 & 544 & 7 & 882 \\
\hshline
\ours & 608 & 66.27 & 11 & - & - & 7 & 18 \\
\end{tabular}
\caption{\textbf{Region \vs grid features} on the VQA 2.0 \testset with accuracy and inference time breakdown measured in milliseconds per image. Our grid features achieve comparable VQA accuracy to region features while being much faster without region feature computation and region selection.}
\label{tab:running_time}
\vspace{-2mm}
\end{table}

\subsection{Test Accuracy and Inference Time}
We now report results on the VQA 2.0 \testset set to quantify the difference in performance between region and grid features. Note that different from previous setups, we use {\trainset}+\valminusminival for training. We report the VQA accuracy and the inference time breakdown in Table~\ref{tab:running_time}. 
Unlike our grid features which directly use convolutional feature maps, region features involve additional operations of region selection and region feature computation. These additional operations take 98.3\% of the total inference time for a region-based model. As a result, the VQA model that takes our grid features as input runs {\bf 48}$\mathbf{\times}$ faster than its counterpart using bottom-up region features.

\begin{figure*}[t]
\centering
\renewcommand{\tabcolsep}{0.25mm}
\newcommand{\FigWidth}{0.12\linewidth}
\newcommand{\FigHeight}{0.076\linewidth}
\newcommand{\question}[1]{\footnotesize\textbf{Q}: #1}
\newcommand{\gtanswer}[1]{\footnotesize\textbf{GT-A}: #1}
\newcommand{\answer}[1]{\footnotesize\textbf{A}#1}
\newcommand{\rightanswer}{\textcolor{green}{\ding{51}}}
\newcommand{\wronganswer}{\textcolor{red}{\ding{55}}}
\begin{tabular}{cccccccc}
    \multicolumn{2}{c}{\question{Which devices do you see?}} & 
    \multicolumn{2}{c}{\question{Has the pizza been eaten?}} & 
    \multicolumn{2}{c}{\question{What color are the curtains?}} & 
    \multicolumn{2}{c}{\question{What is the cat laying on?}} \\
    \multicolumn{2}{c}{\gtanswer{phones}} & 
    \multicolumn{2}{c}{\gtanswer{no}} & 
    \multicolumn{2}{c}{\gtanswer{red and white}} & 
    \multicolumn{2}{c}{\gtanswer{suitcase}} \\
    \answer{(\butd): phones}~\rightanswer &
    \answer{(\ours): phones}~\rightanswer &
    \answer{(\butd): no}~\rightanswer &
    \answer{(\ours): yes}~\wronganswer &
    \answer{(\butd): red}~\wronganswer &
    \answer{(\ours): red and white}~\rightanswer &
    \answer{(\butd): shoes}~\wronganswer &
    \answer{(\ours): shoe}~\wronganswer \\
    \includegraphics[height=\FigHeight]{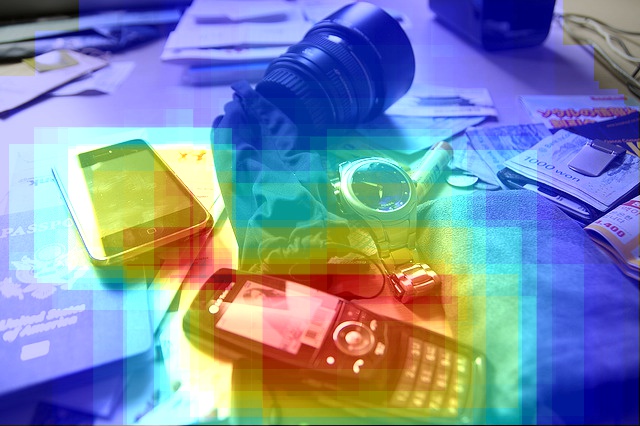} &
    \includegraphics[height=\FigHeight]{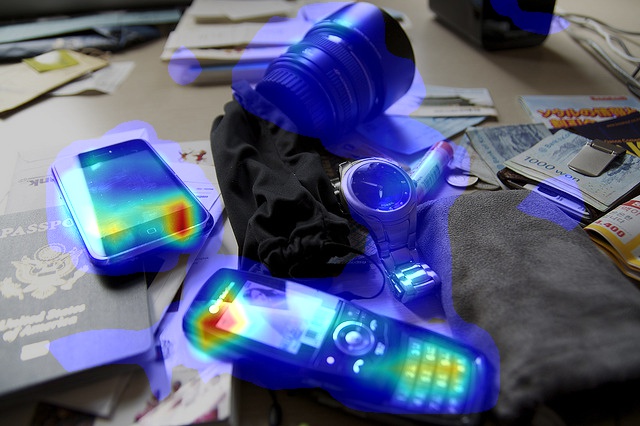} &
    \includegraphics[height=\FigHeight]{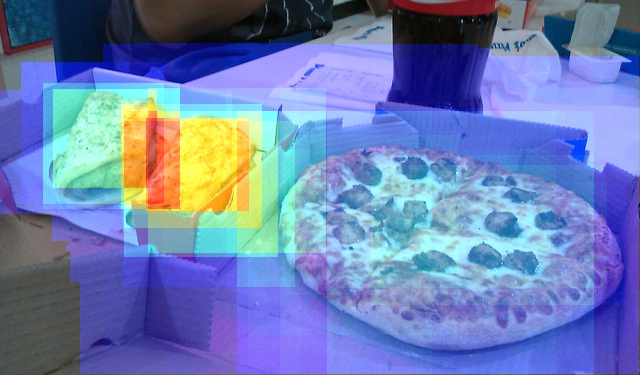} &
    \includegraphics[height=\FigHeight]{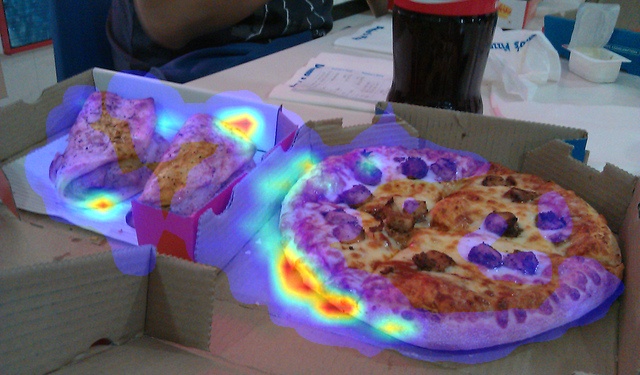} &
    \includegraphics[height=\FigHeight]{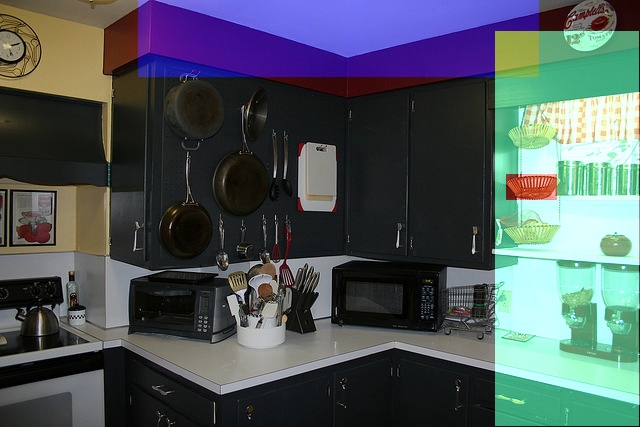} &
    \includegraphics[height=\FigHeight]{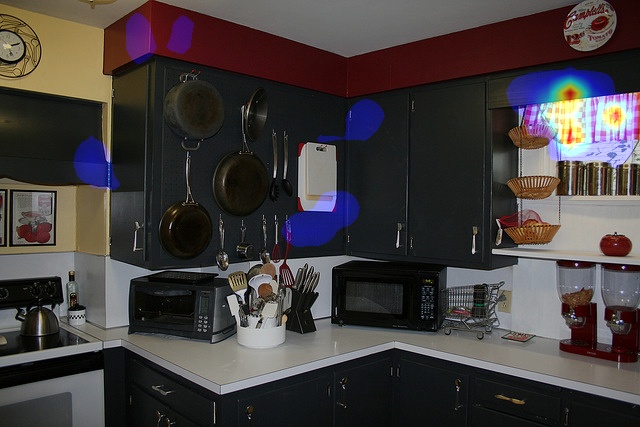} &
    \includegraphics[height=\FigHeight]{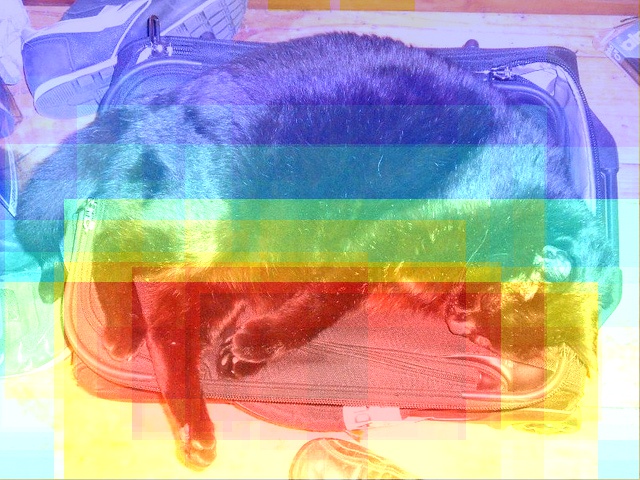} &
    \includegraphics[height=\FigHeight]{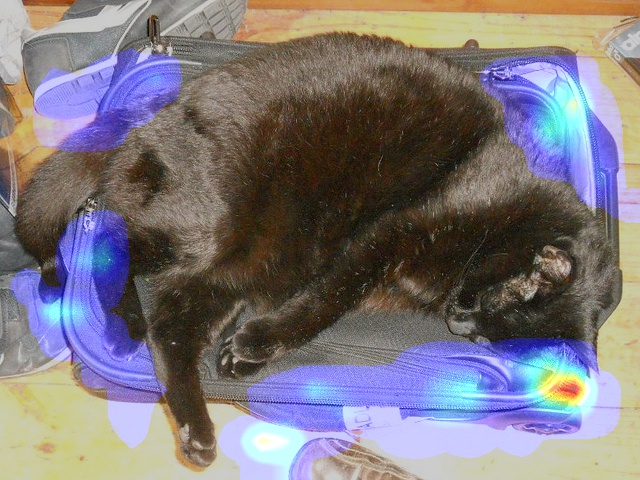} \\
    \multicolumn{2}{c}{\question{Is the plate white?}} & 
    \multicolumn{2}{c}{\question{What breed of dog is this?}} & 
    \multicolumn{2}{c}{\question{What is the person doing?}} & 
    \multicolumn{2}{c}{\question{How many boats do you see?}} \\
    \multicolumn{2}{c}{\gtanswer{yes}} & 
    \multicolumn{2}{c}{\gtanswer{pug}} & 
    \multicolumn{2}{c}{\gtanswer{cutting}} & 
    \multicolumn{2}{c}{\gtanswer{7}} \\
    \answer{(\butd): yes~\rightanswer} &
    \answer{(\ours): yes~\rightanswer} &
    \answer{(\butd): pug}~\rightanswer &
    \answer{(\ours): bulldog}~\wronganswer &
    \answer{(\butd): texting}~\wronganswer &
    \answer{(\ours): cutting}~\rightanswer &
    \answer{(\butd): 5}~\wronganswer &
    \answer{(\ours): 4}~\wronganswer \\
    \includegraphics[height=\FigHeight]{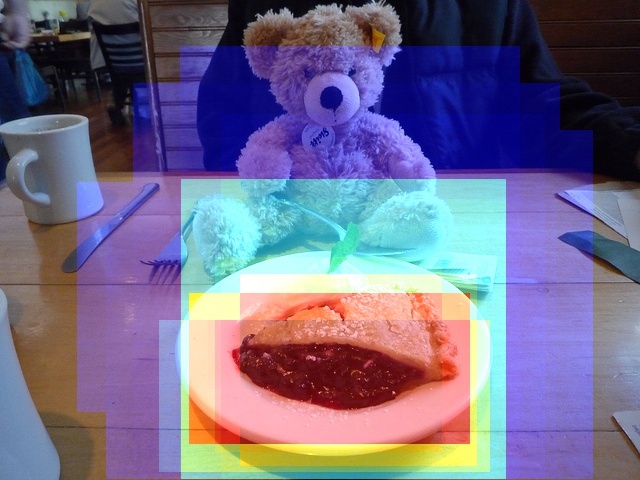} &
    \includegraphics[height=\FigHeight]{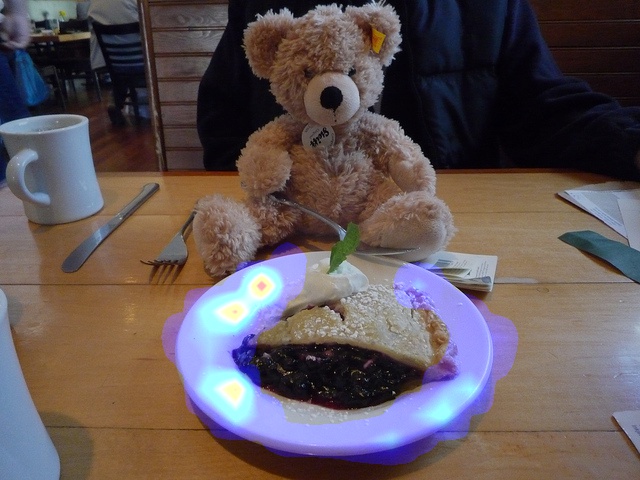} &
    \includegraphics[height=\FigHeight]{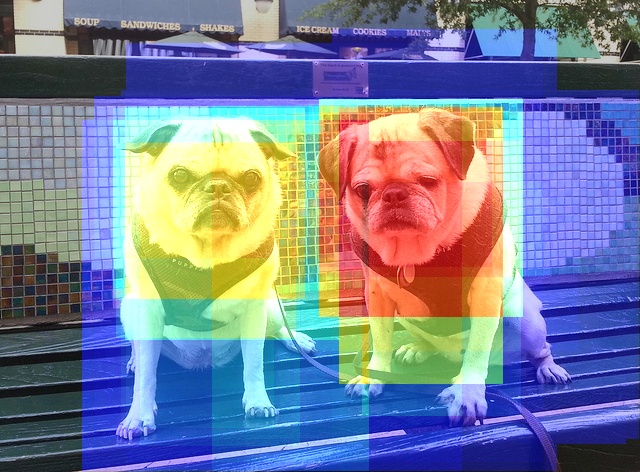} &
    \includegraphics[height=\FigHeight]{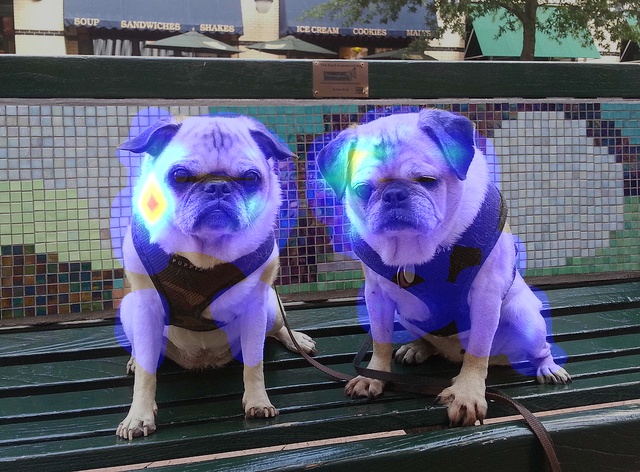} &
    \includegraphics[height=\FigHeight]{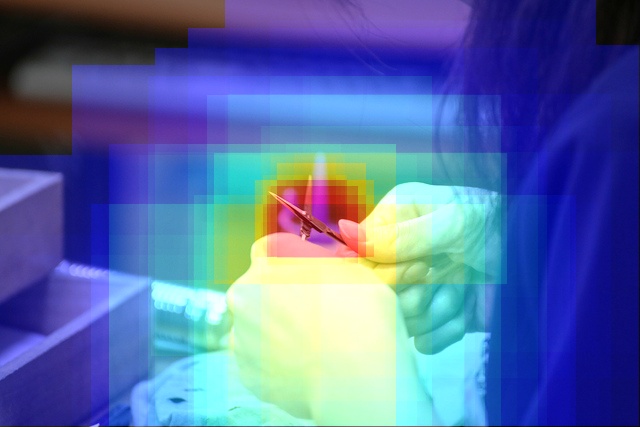} &
    \includegraphics[height=\FigHeight]{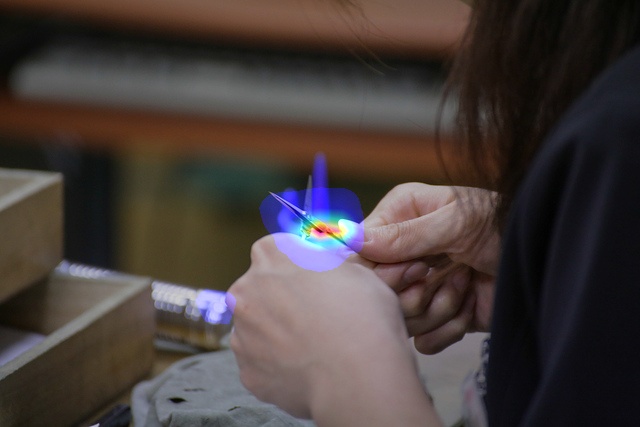} &
    \includegraphics[height=\FigHeight]{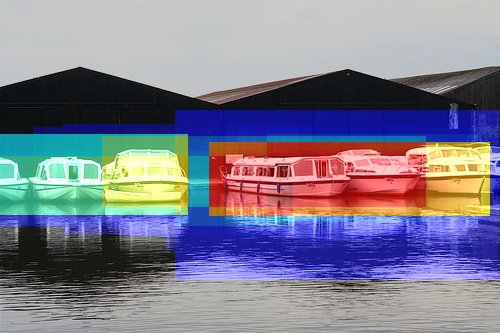} &
    \includegraphics[height=\FigHeight]{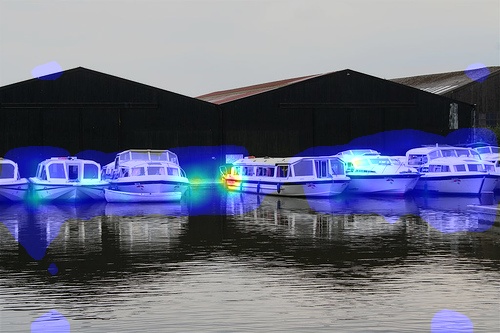} \\
    \multicolumn{2}{c}{(a)} & 
    \multicolumn{2}{c}{(b)} & 
    \multicolumn{2}{c}{(c)} & 
    \multicolumn{2}{c}{(d)} \\
\end{tabular}
\vspace{-4pt}
\caption{\textbf{Visualizations of attention maps overlaid on images} produced by VQA models~\cite{jiang2018pythia}. Source images taken from COCO~\cite{lin2014microsoft} to compare against bottom-up attention~\cite{anderson2018bottom} on VQA 2.0~\cite{goyal2017making}. We show questions (Q), ground-truth answers (GT-A), and side-by-side predictions (attention maps, answers) of region (R) and grid (G) features. \textbf{From left to right}: (a) both region and grid features give correct answers, (b) region features give correct answers but grid features fail, (c) region features fail but grid features give correct answers, and (d) both region and grid features fail. Best viewed in color.}
\label{fig:top_down_attention}
\end{figure*}

\subsection{Qualitative Comparison}
\label{subsec:type_of_questions}
We visualize attention maps over input images from the top-down attention module~\cite{anderson2018bottom}, together with answers from both regions and grids in Fig.~\ref{fig:top_down_attention}. Source images are taken from COCO~\cite{lin2014microsoft} on which VQA 2.0~\cite{goyal2017making} benchmark is built. To obtain the attention map, we propagate the attention value of each region or grid to its corresponding pixels, and then average the attention value for each pixel (normalizing them individually to [0, 1]).
As can be seen, both types of features are able to capture relevant concepts in input images (\eg, snowfield in the top left). Naturally, attention maps of region features tend to cover object-like regions, while for grid features the attention does not necessarily cover the full area the supporting concept (\eg, the snowfield), which can be used to answer the question. However, both features are able to answer visual questions well, suggesting that localization is important, but accurate object detection of individual objects is not crucial for VQA~\cite{goyal2017making}. 

We show failure cases of region and grid features in Fig.~\ref{fig:top_down_attention} (b)(c)(d). In most examples, the models attend to the supporting concepts but still give wrong answers. In the cases where both region and grid features fail, specifically designed modules may be needed (\eg, counting module~\cite{zhang2018learning,trott2018interpretable} in the bottom right example) to answer the question correctly.

\begin{table}[t]
\tablestyle{2.5pt}{1.2}
\centering
\begin{tabular}{c|c|c|c|c}
& & accuracy & pre-training task & input size \\
\shline
\multirow{2}{*}{\ours} & \emph{prev.} & 60.76 & ImageNet~\cite{deng2009imagenet} classification & 448{\x}448 \\
 & \emph{ours} & 64.37 & VG~\cite{krishna2017visual} object+attribute detection & 600{\x}1000 \\
\end{tabular}
\vspace{2mm}
\caption{Comparison between the conventional \textbf{ImageNet pre-trained and our proposed grid features} on the VQA 2.0 \valminusminival set. Besides VQA accuracy, we list two major differences between the two: 1) pre-training task and 2) input image size.}
\label{tab:baseline_results}
\vspace{-2mm}
\end{table}

\section{Why do Our Grid Features Work?}
\label{sec:analysis}
As we mentioned in Sec.~\ref{sec:related_work}, grid features are not new -- in fact, they were widely used in vision and language tasks before the introduction of bottom-up attention features. Compared to the previous attempts at grid features, why do \emph{our} grid features work well? In Table~\ref{tab:baseline_results} we show the performance of grid-based methods (ResNet-50 $C_{5}$ features) for different settings and find that there are two major factors: 1) input image size; 2) pre-training task. We study both these factors next and report results on the \valminusminival set.

\subsection{Factor 1: Input Image Size}
The standard image size used during feature extraction for ImageNet pre-trained models is 448{\x}448~\cite{fukui2016multimodal} discarding the aspect ratio; whereas for VG detection in bottom-up attention~\cite{anderson2018bottom}, the default size is 600{\x}1000 while keeping the aspect ratio intact. Therefore, we experimented with different combinations and reported results for all of them in Table~\ref{tab:effect_image_resolution}. We note that for grid features, a larger input size means more features for the VQA model.

From the table, we find that grid features benefit from larger images as input, indicating this factor is indeed important. However, input size has a different effect for models pre-trained on ImageNet \vs VG. For ImageNet models which are pre-trained on smaller images~\cite{he2016deep}, the performance saturates around 600{\x}1000. Interestingly, the performance of VG models improves with the input size and continues to increase even at 800{\x}1333. We still use 600{\x}1000 for the rest of the paper.

\begin{table}[t]
\tablestyle{5pt}{1.2}
\begin{tabular}{c|c|cc|c|c}
 & \multirow{2}{*}{dataset} & \multicolumn{2}{c|}{input size} & \multirow{2}{*}{\makecell{\# features \\ $N$}} & \multirow{2}{*}{accuracy} \\
 \cline{3-4}
 & & shorter side & longer side & & \\
\shline
\multirow{8}{*}{\ours} & \multirow{4}{*}{\rotatebox[origin=c]{90}{ImageNet}} & 448 & 448 & 196 & 60.76 \\
 & & 448 & 746 & 336 & 61.21 \\
 & & 600 & 1000 & 608 & 61.52 \\
 & & 800 & 1333 & 1050 & 61.52 \\
 \cline{2-6}
 & \multirow{4}{*}{\rotatebox[origin=c]{90}{VG}} & 448 & 448 & 196 & 63.24 \\
 & & 448 & 746 & 336 & 63.81 \\
 & & 600 & 1000 & 608 & 64.37 \\
 & & 800 & 1333 & 1050 & 64.61 \\
\end{tabular}
\caption{\textbf{Impact of input image size} on the VQA 2.0 \valminusminival set. Grid features benefit from larger input image sizes. For an ImageNet pre-trained model, the accuracy saturates around 600{\x}1000 but the VG model makes a better use of larger input image sizes.}
\label{tab:effect_image_resolution}
\vspace{-2mm}
\end{table}

\subsection{Factor 2: Pre-Training Task}
We now study the difference in VQA accuracy due to the pre-training task in the ImageNet (classification) and VG (detection)\footnote{Strictly speaking, VG also uses ImageNet classification for pre-training, because the detector is fine-tuned from a standard ImageNet pre-trained model.}. To understand these differences better, we introduce an additional pre-trained model in each setting. For classification, we include a model trained on YFCC~\cite{thomee2016yfcc100m}, which has 92M images with image tags. For detection, we include a standard model from COCO~\cite{lin2014microsoft} which only has object annotations (no attributes). All models use a ResNet-50 backbone for fair comparison.

The results are shown in Table~\ref{tab:change_task}. In the image classification pre-trained setting, the YFCC model (trained on weak image level tags), performs better than the ImageNet model, possibly because it is trained on two orders of magnitude more data. For detection based pre-training, the VG model (trained with objects and attributes) gives better results than the COCO model. The larger number of categories in VG compared to COCO (1600 \vs 80) or the additional attribute annotations it has are two possible reasons for the improved performance. We study the impact of attributes next.

\begin{table}[t]
\tablestyle{4pt}{1.3}
\begin{tabular}{c|c|c|c|c|c}
& \multicolumn{4}{c|}{pre-train task} & \multirow{2}{*}{accuracy} \\
\cline{2-5}
& setup & dataset & annotation & \#images & \\
\shline
\multirow{4}{*}{\ours} & cls & ImageNet~\cite{deng2009imagenet} & image label & 1.3M & 61.52 \\
& cls & YFCC~\cite{thomee2016yfcc100m} & image tag & 92M & 62.72 \\
\cline{2-6}
& det & COCO~\cite{lin2014microsoft} & object box & 118K & 62.46 \\
& det & VG~\cite{krishna2017visual} & object+attribute & 103K & 64.37 \\
\end{tabular}
\caption{\textbf{Choice of pre-training task.} We explore the impact of the type of pre-training task on the final performance while keeping the input size fixed at 600{\x}1000. Results reported on \valminusminival. We broadly characterize the pre-training tasks into two types - object detection (`det') and image classification (`cls').}
\label{tab:change_task}
\vspace{-2mm}
\end{table}

\begin{figure}
\centering
\includegraphics[width=0.8\linewidth]{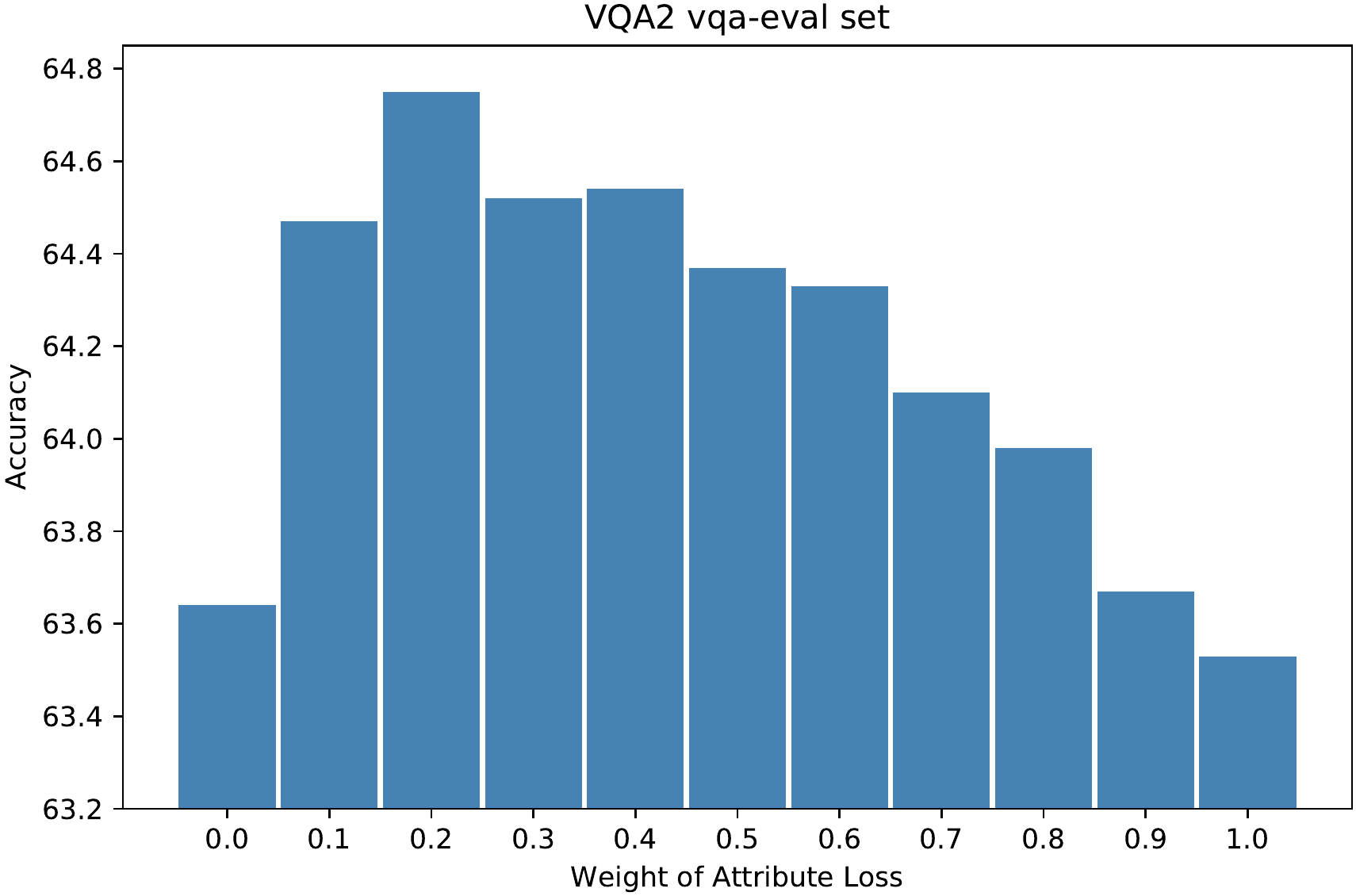}
\vspace{-4pt}
\caption{Analysis on \textbf{attribute loss weights} when pre-training grid features on Visual Genome (VG). All results on VQA 2.0 \valminusminival set. \label{fig:attr_loss_wt}}
\vspace{-6mm}
\end{figure}

\paragraph{Attributes.} Fig.~\ref{fig:attr_loss_wt} shows the impact of the attribute loss weight on VQA accuracy. Setting the attribute loss weight to zero during pre-training on VG, results in a drop in VQA performance. In fact, the VQA accuracy in this case matches the accuracy from a pre-trained COCO model suggesting that attributes in the pre-training task are a major reason for the better performance of VG models. We also note that the grid features consistently outperform the region features for all values of the attribute loss weight.

\section{Generalization of Grid Features}
\label{sec:generalization}
We now study whether our findings about grid features are more broadly applicable to other tasks and models. In this section, we study generalization across: 1) different backbones; 2) different VQA models; 3) different VQA tasks; 4) other tasks. For all the studies, we set the attribute loss weight to 0.2, and compare both the accuracy and speed. For regions we use top $N$=100 ones. Detailed hyper-parameters are in the supplementary material.

\paragraph{Different backbone.} We train Faster R-CNN models with ResNeXt-101-32x8d~\cite{xie2017aggregated} backbone on VG and use the same Pythia setting from Section~\ref{subsec:type_of_questions}. Results on VQA 2.0 \testset split are reported in Table~\ref{tab:generalization:backbone}. We find that our grid features are competitive to the region features even on this more powerful backbone model. Speed-wise, grid features still run substantially faster (23.8{\x}) than region ones.

\paragraph{Different VQA model.} We further test our features obtained from the previous ResNeXt-101 backbone with the state-of-the-art VQA model, MCAN~\cite{yu2019deep} (2019 VQA Challenge winner). We use the open-sourced implementation\footnote{\url{https://github.com/MILVLG/mcan-vqa}} to train the \emph{large} version of the model. The results on VQA 2.0 \testset set are in Table~\ref{tab:generalization:vqa_model}, where our own region features perform better than the results reported in~\cite{yu2019deep} due to stronger backbone. On top of that, our grid features work even \emph{better} than regions, leading to significant improvement over results reported in MCAN~\cite{yu2019deep} (+1.66). This final model reports a state-of-the-art \testfinal result of {\bf 72.71} (single-model performance) for future reference.

\paragraph{Different VQA task.} We use the VizWiz VQA dataset~\cite{gurari2018vizwiz}, which is a real world dataset of pictures taken with cellphones by visually-impaired users. It is more challenging due to poor image quality, conversation-style questions, and unanswerable questions, \etc. Pythia~\cite{jiang2018pythia} model is used (2018 challenge winner). Results on the \testset set of VizWiz are reported in Table~\ref{tab:generalization:vqa_tasks}, where our grid features achieve comparable results to the regions. It is worth pointing out that our grid features run much faster (23{\x}), which provides great potential to be deployed in practice, \eg, on cell phones, to better assist the visually-impaired.

\begin{table*}
\centering
\subfloat[
\label{tab:generalization:backbone}]
{\makebox[0.21\linewidth][c]{
\tablestyle{3pt}{1.2}
\begin{tabular}{c|c|c}
& accuracy & \makecell{time\\(ms)} \\
\shline
Pythia \cite{jiang2018pythia} & 68.31 & - \\
\butd & 68.21 & 929 \\
\hshline
\ours & 67.76 & 39 \\
\end{tabular}
} 
}
\hfill
\subfloat[
\label{tab:generalization:vqa_model}]
{\makebox[0.21\linewidth][c]{
\tablestyle{3pt}{1.2}
\begin{tabular}{c|c|c}
& accuracy & \makecell{time\\(ms)} \\
\shline
MCAN \cite{yu2019deep} & 70.93 & - \\
\butd & 72.01 & 963 \\
\hshline
\ours & 72.59 & 72 \\
\end{tabular}
} 
}
\hfill
\subfloat[
\label{tab:generalization:vqa_tasks}]
{\makebox[0.21\linewidth][c]{
\tablestyle{3pt}{1.2}
\begin{tabular}{c|c|c}
& accuracy & \makecell{time\\(ms)} \\
\shline
Pythia \cite{jiang2018pythia} & 54.22 & - \\
\butd & 54.28 & 874 \\
\hshline
\ours & 54.17 & 38 \\
\end{tabular}
} 
}
\hfill
\subfloat[ 
\label{tab:generalization:other_tasks}]
{\makebox[0.34\linewidth][c]{
\tablestyle{3pt}{1.2}
\begin{tabular}{c|cccc|c}
& B4 & M & C & S & \makecell{time\\(ms)} \\
\shline
BUTD \cite{anderson2018bottom} & 36.2 & 27.0 & 113.5 & 20.3 & - \\
\butd & 36.2 & 27.7 & 113.9 & 20.8 & 1101 \\
\hshline
\ours & 36.4 & 27.4 & 113.8 & 20.7 & 240 \\
\end{tabular}
} 
}
\vspace{-6pt}
\caption{\textbf{Generalizations of grid features}. From left to right: (a) Different \textbf{backbone.} We use a ResNeXt-101-32x8d instead of a ResNet-50 as the backbone. (b) Different \textbf{VQA model}. We use MCAN~\cite{yu2019deep} implementation which is the state-of-the-art VQA model. (c) Accuracy on \textbf{VizWiz} using the same VQA models~\cite{jiang2018pythia}. (d) \textbf{Image captioning}  on COCO Karpathy test split. Abbreviations: BLEU4 (B4), METEOR (M), CIDEr (C), and SPICE (S). Our grid features generalize well by achieving results at-par with bottom-up region features while being significantly faster.}
\vspace{-2mm}
\end{table*}

\paragraph{Image captioning.} We train the bottom-up attention model~\cite{anderson2018bottom} implemented in Pythia~\cite{jiang2018pythia} taking our features as input for image captioning on COCO~\cite{chen2015microsoft}. No CIDEr~\cite{vedantam2015cider} optimization~\cite{anderson2018bottom} is used for fair comparison. Quantitative results on the test set of Karpathy split~\cite{karpathy2015deep} are reported in Table~\ref{tab:generalization:other_tasks}. We use standard evaluation metrics including BLEU4~\cite{papineni2002bleu}, METEOR~\cite{lavie2007meteor}, CIDEr, and SPICE~\cite{anderson2016spice}. Similar to the VQA task, our grid features achieve comparable results to bottom-up region ones for image captioning while being significantly faster.

\begin{table}[t]
\tablestyle{6pt}{1.1}
\begin{tabular}{c|c|c|c|c|c}
\multicolumn{2}{c|}{pre-train task} & \multirow{2}{*}{e2e} & \multirow{2}{*}{\makecell{PPM\\\cite{zhao2017pyramid}}} & \multirow{2}{*}{accuracy} & \multirow{2}{*}{$\Delta$} \\
\cline{1-2}
dataset & \makecell{region\\annotations?} & & & & \\
\shline
 \multirow{3}{*}{\makecell{VG~\cite{krishna2017visual}}} & \multirow{3}{*}{\large \cmark} & & & \underline{66.27} & - \\
&  & \cmark & & 66.47 & \emph{0.20} \\
&  & \cmark & \cmark & 66.74 & \emph{0.47}  \\
\hline
\multirow{3}{*}{\makecell{ImageNet~\cite{deng2009imagenet}}} & \multirow{3}{*}{\large \xmark} & & & \underline{63.21} & - \\
&  & \cmark & & 64.98 & \emph{1.77} \\
&  & \cmark & \cmark & 65.97 & \emph{2.76} \\
\hline
\multirow{3}{*}{\makecell{YFCC~\cite{thomee2016yfcc100m}}} & \multirow{3}{*}{\large \xmark} & & & \underline{65.04} & - \\
& & \cmark & & 65.35 & \emph{0.31} \\
&  & \cmark & \cmark & 66.61 & \emph{1.57} \\
\end{tabular}
\caption{Results of \textbf{end-to-end trained VQA} models with grid features on the VQA 2.0 \testset set. End-to-end learning boosts accuracy for all models and more for ones trained on ImageNet and YFCC. Adding PPM~\cite{zhao2017pyramid} further improves accuracy.}
\label{tab:e2e_vqa}
\vspace{-2mm}
\end{table}

\section{Towards End-to-end VQA}
\label{sec:e2e_vqa}
Although pre-training on VG, ImageNet, or YFCC provides useful feature representations for VQA, there are still potential domain shifts between the pre-training tasks and the downstream tasks. For example, YFCC contains a lot of outdoor images~\cite{thomee2016yfcc100m}, which are not present in the VQA dataset. Instead of using pre-computed fixed feature representations, \emph{end-to-end} training, where the initial feature representations will be fine-tuned, provides a natural solution to reducing such domain gaps. Empowered by the dramatic simplification of grid features for the VQA pipeline, we take an initial step towards this goal.

\paragraph{Training details.} We adopt the 22K learning rate schedule~\cite{jiang2018pythia} to train both the ResNet-50 model and the Pythia VQA model jointly, with errors from the answering accuracy \emph{directly} back-propagated to the grid convolutional feature maps. We fix the first two residual blocks and fine-tune the rest of the model. Since the visual representations are computed online (not stored on disk), it allows us to perform data augmentation including color jitter and affine transformation over the input images to reduce chance of over-fitting. For more details see supplementary material.

\paragraph{Results.} We experiment with three models pre-trained on VG, ImageNet, and YFCC. Note that while VG uses \emph{region}-level annotations, both ImageNet and YFCC only use \emph{image}-level ones (human labels or noisy image tags). As can be seen from Table~\ref{tab:e2e_vqa}, end-to-end training (denoted as `e2e') can boost accuracy for all three pre-trained models, with the biggest improvements for ImageNet models.

\paragraph{Flexible network design.} As we now have the ability to train our models end-to-end in a simple manner, it allows us to introduce more flexible architectural designs for vision and language tasks~\cite{lu2019vilbert}. Specifically, on top of the grid features from the ResNet-50 model, we add a Pyramid Pooling Module (PPM, a component widely used for semantic segmentation; details in supplementary material)~\cite{zhao2017pyramid,xiao2018unified} to aggregate visual information from grid features of different spatial resolutions. After adding this module to different pre-trained models (Table~\ref{tab:e2e_vqa}, `PPM'), the VQA accuracy can be further improved. Remarkably, for ImageNet and YFCC pre-trained models, a combination of end-to-end training and PPM results in close or even \emph{better} performance than a VG pre-trained model using pre-computed region features. This result is particularly desirable as it indicates good VQA accuracy can be achieved even with \emph{zero} use of explicit region (bounding box) annotations.

\section{Conclusion}
\label{sec:discussions}
In this paper, we revisit grid features as an alternative to the widely used bottom-up region features~\cite{anderson2018bottom} for vision and language tasks. We show they can in fact achieve on-par results in terms of accuracy over different VQA tasks and models and even on captioning. 
As a result of skipping the computationally expensive region-related bottlenecks in the pipeline, we see remarkable speed-ups -- often more than an order of magnitude -- to the existing systems that rely on regions. Our experiments show that rather than the `format' of features (region \vs grids), the semantic content that features represent is more critical for their effectiveness. Such effective representation, per our experiment, can be achieved either by pre-training on object and attribute datasets such as VG, or more importantly, by \emph{end-to-end} training of grid features directly for the end-task. Note that while easy with grid-features, end-to-end training is not trivial with regions.
Even with limited exploration in this direction, we already find that given a more flexible design space, grid features pre-trained without \emph{any} region-level annotations can in fact achieve strong performance on VQA. While we are aware that for tasks like referring expressions~\cite{kazemzadeh2014referitgame} where the output itself is a region, modeling region is likely unavoidable, we hope our grid features can potentially offer new perspectives for vision and language research in general. 

\vspace{2mm}
\footnotesize
\paragraph{Acknowledgements.} We would like to thank Larry Zitnick, Duy-Kien Nguyen, Devi Parikh, and Amanpreet Singh for helpful discussions. ELM acknowledges support from
AFRL and DARPA (\#FA8750- 18-2-0126). The U.S. government is authorized
to reproduce and distribute reprints for government purposes notwithstanding any
copyright notation thereon. The views and conclusions contained herein
are those of the authors and should not be interpreted as necessarily representing the official policies or endorsements, either expressed or implied, of the AFRL and DARPA or the U.S. government.

\normalsize
\appendixtitleon
\appendixtitletocon
\begin{appendices}

\onecolumn

\section{Details of Hyperparameters}
\label{sec:hyper_param}

\begin{table*}[h]
\tablestyle{5pt}{1.2}
\begin{tabular}{c|c|c|c|c|c|c|c|c}
model & dataset & optimizer & \# iterations & \makecell{batch\\size} & initial lr & lr decay & lr schedule & \makecell{gradient\\clip} \\
\shline
Faster R-CNN & VG/COCO & SGD & 90K & 16 & 0.02 & 0.1 & [60K, 80K] & -\\
\cite{jiang2018pythia} & VQA 2.0,~\trainonly & Adamax~\cite{kingma2015adam} & 12K & 512 & 0.01 & 0.1 & [5K, 7K, 9K, 11K] & 0.25 \\
\cite{jiang2018pythia} & VQA 2.0,~\trainonly + \valminusminival & Adamax & 22K & 512& 0.01 & 0.1 & [15K, 18K, 20K, 21K] & 0.25 \\
MCAN~\cite{yu2019deep} & VQA 2.0~\trainset + VG & Adam & 234K\footnotemark & 64 & 5e-5 & 0.02 & [180K, 216K] & - \\
\cite{jiang2018pythia} & VizWiz & Adamax & 24K & 128 & 0.005 & 0.01 & [14K] & 0.25 \\
\cite{anderson2018bottom}\footnotemark & COCO Karpathy split & Adamax & 50K & 256 & 0.002 & 0.1 & [15K, 25K, 35K, 45K] & 0.25 \\
\hline
e2e~\cite{jiang2018pythia} & VQA 2.0,~\trainonly +\valminusminival & Adamax & 22K & 512& 0.002 & 0.1 & [15K, 18K, 20K, 21K] & 1 \\
\end{tabular}
\caption{\textbf{Summary of hyperparameters}. We follow the default setting for most of the models. For the image captioning model~\cite{anderson2018bottom,jiang2018pythia}, the default initial learning rate is 0.01. We found 0.002 leads to slightly better results. For the end-to-end trained Pythia (e2e Pythia in the last row), we use initial learning rate of 0.002 and a larger value of 1 for the gradient clip when fine-tuning the ResNet model for feature extraction.}
\label{tab:hyper_param}
\end{table*}
\footnotetext[1]{In the MCAN paper, the model is trained for 13 epochs, where each epoch contains 17,967 iterations.}
\footnotetext[2]{We use the implementation provided in~\cite{jiang2018pythia}.}

Hyper-parameters of different models are summarized in Table~\ref{tab:hyper_param}. For the SGD optimizer, the momentum is 0.9 and weight decay is 0.0001. For the Adamax optimizer, $\beta_1$ and $\beta_2$ are 0.9 and 0.999, respectively. No weight decay is used. For the Adam optimizer used in MCAN~\cite{yu2019deep}, $\beta_1$ and $\beta_2$ are 0.9 and 0.98, respectively. No weight decay is used. 

We follow the default setting of hyperparameters for most of models. For the image captioning model~\cite{anderson2018bottom,jiang2018pythia}, the default initial learning rate is 0.01. We found 0.002 leads to slightly better results. For the end-to-end trained Pythia (e2e Pythia in the last row), we use an initial learning rate of 0.002 and a larger value of 1 for the gradient clip when fine-tuning the ResNet model for feature extraction.

\newcommand{\fcsix}{\texttt{fc6}\xspace}
\newcommand{\fcseven}{\texttt{fc7}\xspace}

\section{Region Features from FPN}
\label{sec:fpn}
In the Pythia implementation~\cite{jiang2018pythia}, of bottom-up attention~\cite{anderson2018bottom}, a Feature Pyramid Network (FPN) model~\cite{lin2017feature} is used to compute region features. This is different from the original Faster R-CNN model~\cite{ren2015faster} used, and it is commonly believed that FPN can offer \emph{better} object detection quality. Therefore, to reach a more solid conclusion, in this appendix we show extended results from the main paper to compare our grid features with FPN region features. The FPN model uses an entire ResNet model as the backbone, where the multi-scale feature maps of different blocks of the ResNet model are fused in a feature pyramid. Two randomly initialized fully-connect layers (denoted as~\fcsix and~\fcseven for simplicity) are added to predict object category, bounding box regression offsets, and attribute labels for each bounding box proposal. We follow the strategy used in~\cite{jiang2018pythia} to compute region features. Specifically, we use the output of the \fcsix layer as input to a VQA or image captioning model, where the \fcseven layer is also used and \emph{fine-tuned} during VQA training.

Accuracy on the VQA 2.0~\testset set and breakdown inference time of the FPN model, using a ResNet50 as the backbone, are summarized in Table~\ref{tab:fpn_running_time}. Different from the trend observed in object detection~\cite{lin2017feature}, we find the FPN model, when used to provide region features for VQA, does not show clear advantage over the original C4 model~\cite{anderson2018bottom}, which in turn gives on-par results to our grid features. Speed-wise, despite the lighter pre-region computation, we find the region-related steps with FPN are still very expensive, and the efficiency advantage of our grid features is even more significant.

We also test the top 100 ($N{=}100$) regions using different backbones, VQA models, VQA tasks, and image captioning task, as we have done in Section 6 in the paper. Results are reported in Table~\ref{tab:fpn_generalization:backbone},~\ref{tab:fpn_generalization:vqa_model},~\ref{tab:fpn_generalization:vqa_tasks}, and~\ref{tab:fpn_generalization:other_tasks}. For the accuracy on the VQA 2.0~\testset set and VizWiz, the FPN model's accuracy is lower than the results reported in~\cite{jiang2018pythia}, because grid features (from an ImageNet pre-trained ResNet-152~\cite{he2016deep} model) are used in \emph{addition} to the region features~\cite{jiang2018pythia}. Using the MCAN model~\cite{yu2019deep}, the FPN model achieves better results than reported in~\cite{yu2019deep} but still performs worse than C4 and our grid features.

\begin{table}[t]
\tablestyle{6pt}{1.2}
\begin{tabular}{c|c|c|c|c|c|c|c}
 & \multirow{3}{*}{\makecell{\# features \\ ($N$)}} & \multirow{3}{*}{\makecell{\testset\\accuracy}} & \multicolumn{5}{c}{inference time breakdown (ms)} \\
 \cline{4-8}
 & & & \makecell{shared\\conv.} & \makecell{region\\feat. comp.} & \makecell{region\\selection} & VQA & total \\
\shline
\multirow{2}{*}{\butd} & 100 & 66.13 & 9 & 326 & 548 & 6 & 889 \\
& 608 & 66.22 & 9 & 322 & 544 & 7 & 882 \\
\hshline
\multirow{2}{*}{\makecell{\butd\\w/ FPN}} & 100 & 66.01 & 11 & 311 & 690 & 5 & 1017 \\
& 608 & 66.36 & 12 & 323 & 690 & 7 & 1032 \\
\hshline
\ours & 608 & 66.27 & 11 & - & - & 7 & 18 \\
\end{tabular}
\caption{This table extends Table 2 in the main paper for {\bf speed and accuracy comparisons} with added rows for region features with FPN. Results are reported on VQA 2.0 \testset with accuracy and inference time breakdown measured in milliseconds per image. Despite the advantages which FPN features have that 1) pools features from higher-resolution feature maps; and 2) fine-tunes the \fcseven layer~\cite{jiang2018pythia} when training VQA; our grid features achieve comparable VQA accuracy to all region features and are much faster.}
\label{tab:fpn_running_time}
\vspace{-1mm}
\end{table}

\begin{table}
\centering
\subfloat[
\label{tab:fpn_generalization:backbone}]
{\makebox[0.45\linewidth][c]{
\tablestyle{6pt}{1.2}
\begin{tabular}{c|c|c|c|c|c}
& \multicolumn{4}{c|}{VQA 2.0 accuracy} & \multirow{2}{*}{\makecell{time\\(ms)}} \\
\cline{2-5}
& Yes/No & Number & Other & Overall & \\
\shline
\cite{jiang2018pythia} & - & - & - & 68.31 & - \\
\butd & 84.73 &	46.88 &	58.98 & 68.21 & 929 \\
\butd, w/ FPN & 83.88 &	45.13 & 58.12 & 67.26 & 1069 \\
\hshline
\ours & 84.13 & 45.98 &	58.76 &	67.76 & 39 \\
\end{tabular}
} 
}
\hfill
\subfloat[
\label{tab:fpn_generalization:vqa_model}]
{\makebox[0.45\linewidth][c]{
\tablestyle{6pt}{1.2}
\begin{tabular}{c|c|c|c|c|c}
& \multicolumn{4}{c|}{VQA 2.0 accuracy} & \multirow{2}{*}{\makecell{time\\(ms)}} \\
\cline{2-5}
& Yes/No & Number & Other & Overall & \\
\shline
\cite{yu2019deep} & 87.39 & 52.78 &	60.98 & 70.93 & - \\
\butd & 88.19 &	54.38 &	62.19 & 72.01 & 963 \\
\butd, w/ FPN & 87.77 &	54.72 &	62.16 &	71.87 & 1100 \\
\hshline
\ours & 88.46 &	55.68 &	62.85 & 72.59 & 72 \\
\end{tabular}
} 
}
\\
\subfloat[
\label{tab:fpn_generalization:vqa_tasks}]
{\makebox[0.45\linewidth][c]{
\tablestyle{3pt}{1.2}
\begin{tabular}{c|c|c|c|c|c|c}
& \multicolumn{5}{c|}{VizWiz accuracy} & \multirow{2}{*}{\makecell{time\\(ms)}} \\
\cline{2-6}
& Yes/No & Number & Other & Un. Ans. & Overall & \\
\shline
\cite{jiang2018pythia} &- & - & - & - & 54.22 & - \\
\butd & 73.17 &	28.89 &	83.63 &	35.62 & 54.28 & 874 \\
\makecell{\butd, w/ FPN} & 73.00 &	27.11 &	82.02 &	33.59 &	52.50 & 1051\\
\hshline
\ours & 75.17 &	24.89 &	83.68 &	35.35 & 54.17 & 38 \\
\end{tabular}
} 
}
\hfill
\subfloat[ 
\label{tab:fpn_generalization:other_tasks}]
{\makebox[0.5\linewidth][c]{
\tablestyle{4pt}{1.2}
\begin{tabular}{c|cccccccc|c}
& B4 & B3 &	B2 & B1 & RL & M & C & S & \makecell{time\\(ms)} \\
\shline
\cite{anderson2018bottom} & 36.2 & - & - & 77.2 & 56.4 & 27.0 & 113.5 & 20.3 & - \\
\butd &	36.2 & 46.8 & 60.4 & 76.4 & 56.5 & 27.7 & 113.9 & 20.8 & 1101 \\
\makecell{\butd, w/ FPN} &	35.7 & 46.5 & 60.3 & 76.6 &	56.4 & 27.5  & 113.1 & 20.6 & 1099 \\
\hshline
\ours &	36.4 & 47.3 & 61.1 & 76.7 &	56.6 & 27.4 & 113.8 & 20.7 & 240 \\
\end{tabular}
} 
}
\caption{This table extends Table 6 in the main paper for {\bf generalization experiments}. From left to right: (a) Different \emph{backbone.} We use a ResNeXt-101-32x8d instead of a ResNet-50 as the backbone. (b) Different \emph{VQA model}. We use MCAN~\cite{yu2019deep} implementation which is the state-of-the-art VQA model. (c) Accuracy on \emph{VizWiz} using the same VQA models~\cite{jiang2018pythia}. (d) \emph{Image captioning}  on COCO Karpathy test split. Abbreviations: BLEU4 (B4), BLEU3 (B3), BLEU2 (B2), BLEU1 (B1), ROUGE\_L (RL), METEOR (M), CIDEr (C), and SPICE (S). Our grid features generalize well by achieving results at-par with bottom-up region features while being significantly faster. }
\vspace{-1mm}
\end{table}

\section{Details of PPM}
\label{sec:ppm}
\begin{figure*}
\centering
\includegraphics[width=0.95\linewidth]{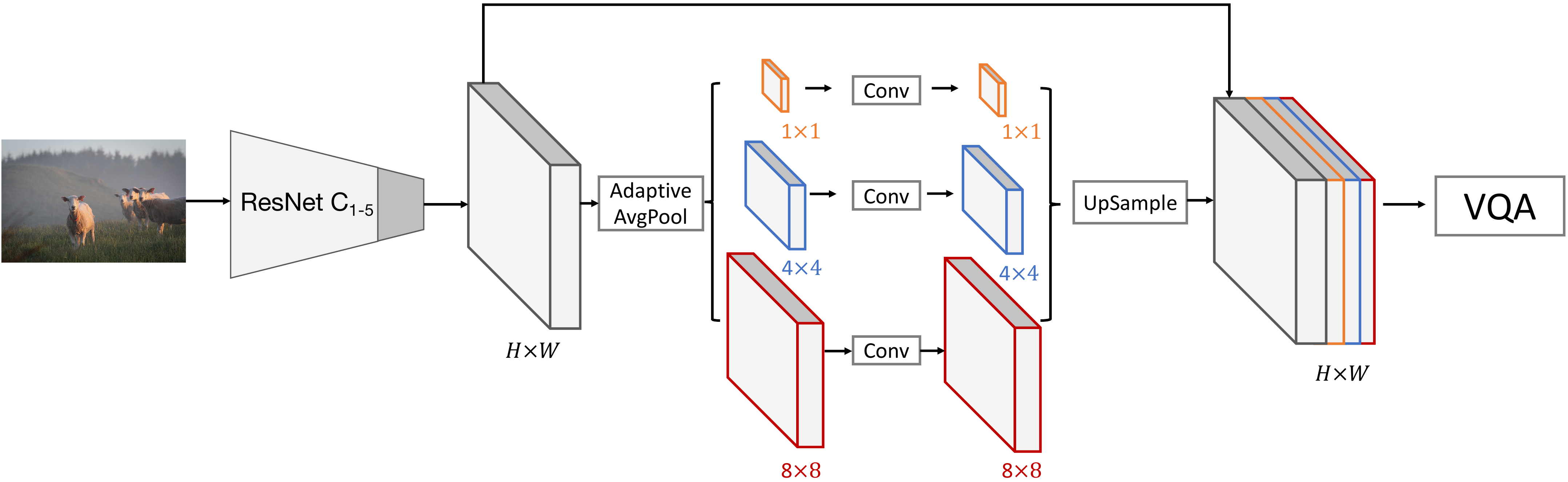}
\caption{Illustration of \textbf{PPM (Pyramid Pooling Module)}~\cite{zhao2017pyramid} experimented in the end-to-end model for VQA. See Section \ref{sec:ppm} for details.}
\label{fig:ppm}
\end{figure*}

In Section 7 of the main paper, we introduce end-to-end training of the Pythia VQA model~\cite{jiang2018pythia} with PPM (Pyramid Pooling Module)~\cite{zhao2017pyramid}.
A detailed illustration of this module is provided in Fig.~\ref{fig:ppm}. Given a grid convolution feature map from a ResNet model, adaptive average pooling operations are performed at three different spatial resolutions: 1{\x}1, 4{\x}4, and 8{\x}8. Three separate convolution layers (followed by batch normalization and ReLU) are added, where the kernel sizes are all set to 1 and output dimensions are all 512. Finally, the original grid feature map is concatenated together with the three ones obtained from PPM as the input for VQA.

\end{appendices}

\newpage

\small
\twocolumn
\bibliographystyle{ieee_fullname}
\bibliography{egbib}

\end{document}